\documentclass[10pt,a4paper]{article}
\usepackage{lrec}
\usepackage{graphicx}
\usepackage{tabularx}
\usepackage{soul}
\usepackage{epstopdf}
\usepackage[latin1]{inputenc}

\usepackage{hyperref}
\usepackage{xstring}
\usepackage{authblk}

\usepackage{times}
\usepackage{booktabs}
\usepackage{latexsym}
\usepackage{caption}
\usepackage{subcaption}
\usepackage{graphicx}
\usepackage{authblk}

\usepackage{multicol}
\usepackage{lipsum}
\usepackage{mwe}

\usepackage{cellspace}
\usepackage[export]{adjustbox}
\newcommand\cincludegraphics[2][]{\raisebox{-0.3\height}{\includegraphics[#1]{#2}}}

\usepackage{booktabs}

\newcommand\mybar{\kern1pt\rule[-\dp\strutbox]{.8pt}{\baselineskip}\kern1pt}

\title{r/Fakeddit:\\ A New Multimodal Benchmark Dataset for\\ Fine-grained Fake News Detection}

\name{Kai Nakamura*\thanks{* Equal Contribution.}\P, Sharon Levy*\S, William Yang Wang\S}

\address{\P Laguna Blanca School \\  \S University of California, Santa Barbara \\
         kai.nakamura42@gmail.com,
         \{sharonlevy, william\}@cs.ucsb.edu\\}

\abstract{
Fake news has altered society in negative ways in politics and culture. It has adversely affected both online social network systems as well as offline communities and conversations. Using automatic machine learning classification models is an efficient way to combat the widespread dissemination of fake news. However, a lack of effective, comprehensive datasets has been a problem for fake news research and detection model development. Prior fake news datasets do not provide multimodal text and image data, metadata, comment data, and fine-grained fake news categorization at the scale and breadth of our dataset. We present Fakeddit, a novel multimodal dataset consisting of over 1 million samples from multiple categories of fake news. After being processed through several stages of review, the samples are labeled according to 2-way, 3-way, and 6-way classification categories through distant supervision. We construct hybrid text+image models and perform extensive experiments for multiple variations of classification, demonstrating the importance of the novel aspect of multimodality and fine-grained classification unique to Fakeddit.  
 \\ \newline  }

\begin{document}

\maketitleabstract

\section{Introduction}

Within our progressively digitized society, the spread of fake news and disinformation has enlarged in journalism, news reporting, social media, and other forms of online information consumption. False information from these sources, in turn, has caused many problems such as spurring irrational fears during medical outbreaks like Ebola\footnote{https://www.pbs.org/newshour/science/real-consequences-fake-news-stories-brain-cant-ignore}. The dissemination and consequences of fake news are exacerbating due to the rise of popular social media applications and other online sources with inadequate fact-checking or third-party filtering, enabling any individual to broadcast fake news easily and at a large scale \cite{10.1257/jep.31.2.211}. Though steps have been taken to detect and eliminate fake news, it still poses a dire threat to society \cite{facebook}. According to a Pew Research Center report\footnote{https://www.journalism.org/2019/06/05/many-americans-say-made-up-news-is-a-critical-problem-that-needs-to-be-fixed/}, 50\% of Americans view fake news as a critical problem, placing it above violent crime. In addition, the report found that 68\% of Americans view fake news as having a significant impact on their confidence of the government and 54\% viewed it as having a large impact in their trust in one another. As such, research in the area of fake news detection is of high importance for society.

\begin{figure*}
  \centering
  \includegraphics[width=\linewidth]{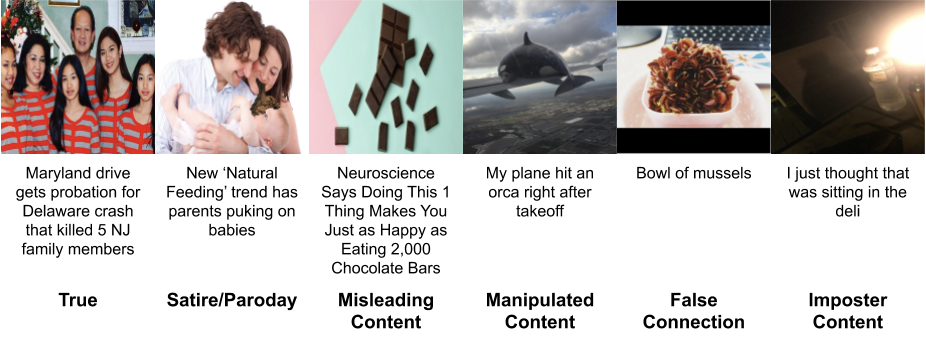}
  \caption{Dataset examples with 6-way classification labels.}\label{fig:dataset}
\end{figure*}

To build a fake news detection model, one must obtain sizable and diverse training data. Within this area of research, there are several existing published datasets. However, they have many constraints: limited size, modality, and granularity. Most conventional fake news research and datasets such as LIAR \cite{wang-2017-liar} and Some-Like-It-Hoax \cite{DBLP:journals/corr/TacchiniBVMA17} solely focus on text data. However, online information today is also consumed through multimedia sources including images, which often supplement the text. In addition, many datasets are small in size and variation. For example, Abu Salem et al. \shortcite{fa-kes} aims to increase the diversity of fake news by covering news that goes beyond the scope of conventional American political news. However, it suffers the problem of only consisting of less than 1000 samples, limiting its extent to which it can contribute to fake news research. Moreover, many conventional datasets label their data binarily (true and false). However, fake news can be categorized into many different types \cite{wardle}. These problems significantly affect the quality of fake news research and detection. 

We overcome these limitations posed by conventional datasets through the dataset we propose: Fakeddit\footnote{\url{https://github.com/entitize/fakeddit}}, a novel multimodal fake news detection dataset consisting of over 1 million samples with 2-way, 3-way, and 6-way classification labels, along with comment data and metadata. We sourced our data from multiple subreddits from Reddit\footnote{http://reddit.com/}. Our dataset will expand fake news research into the multimodal space and allow researchers to develop stronger, more generalized, fine-grained fake news detection systems. We provide examples from our dataset in Figure \ref{fig:dataset}.

Our contributions to the study of fake news detection are:

\begin{itemize}
  \item[$\bullet$] We create a large-scale multimodal fake news dataset consisting of over 1 million samples containing text, image, metadata, and comments data from a highly diverse set of resources. 
  \item[$\bullet$] Each data sample consists of multiple labels, allowing users to utilize the dataset for 2-way, 3-way, and 6-way classification. This enables both high-level and fine-grained fake news classification. Samples are also thoroughly refined through multiple steps of quality assurance. 
  \item[$\bullet$] We evaluate our dataset through text, image, and text+image modes with neural network architectures that integrate both the image and text data. We run experiments for several types of baseline models, providing a comprehensive overview of classification results and demonstrating the significance of multimodality present in Fakeddit.
\end{itemize}

\begin{table*}[t]
\centering
\small
\makebox[\linewidth]{
\begin{tabular}{{l}|{l}|{l}|{l}|{l}|{l}}
\toprule
Dataset &Size (\# of samples)& \# of Classes  & Modality & Source & Data Category   \\
\hline
LIAR & 12,836 & 6 & text & Politifact & political\\
\hline
FEVER & 185,445 & 3 & text & Wikipedia & variety\\
\hline
BUZZFEEDNEWS & 2,282 &4 & text & Facebook & political\\
\hline
BUZZFACE & 2,263 &4 & text &Facebook&political\\
\hline
some-like-it-hoax & 15,500  &2 & text & Facebook& scientific/conspiracy\\
\hline
PHEME & 330 &2 & text & Twitter & variety\\
\hline
CREDBANK & 60,000,000 &5 & text & Twitter & variety\\
\hline
Breaking! & 700 & 2,3 & text & BS Detector & political\\
\hline
NELA-GT-2018 & 713,000 &8 IA &text&194 news outlets&variety\\
\hline
FAKENEWSNET & 602,659 &2 & text & Twitter & political/celebrity \\
\hline
FakeNewsCorpus & 9,400,000 &10 & text & Opensources.co & variety\\
\hline
FA-KES & 804 &2 & text & 15 news outlets &Syrian war\\
\hline
Image Manipulation & 48 &2 & image & self-taken & variety\\
\hline
Fauxtography & 1,233 & 2 & text, image & Snopes, Reuters& variety \\
\hline
image-verification-corpus & 17,806 &2&text, image&Twitter&variety \\
\hline
The PS-Battles Dataset & 102,028 & 2 & image & Reddit & manipulated content \\
\hline
\textbf{Fakeddit (ours)} & \textbf{1,063,106} &\textbf{2,3,6} & \textbf{text, image} & \textbf{Reddit} & \textbf{variety} \\
\bottomrule
 \end{tabular}
 }
\caption{ Comparison of various fake news detection datasets. IA: Individual assessments. }\label{tab:dataset}
\end{table*}

\section{Related Work}
A variety of datasets for fake news detection have been published in recent years. These are listed in Table \ref{tab:dataset}, along with their specific characteristics. 
\subsection{Text Datasets}
When comparing fake news datasets, a few trends can be seen. Most of the datasets are small in size, which can be ineffective for current machine learning models that require large quantities of training data. Only four datasets contain over half a million samples, with CREDBANK~\cite{mitra2015credbank} and FakeNewsCorpus\footnote{https://github.com/several27/FakeNewsCorpus} being the largest, both containing millions of samples. In addition, many of the datasets separate their data into a small number of classes, such as fake vs. true. Datasets such as NELA-GT-2018~\cite{NELA}, LIAR~\cite{wang-2017-liar}, and FakeNewsCorpus provide more fine-grained labels. While some datasets include data from a variety of categories \cite{zubiaga2016analysing}, many contain data from specific areas, such as politics and celebrity gossip~\cite{DBLP:journals/corr/TacchiniBVMA17,pathak-srihari-2019-breaking,shu2018fakenewsnet,fa-kes,santia2018buzzface}\footnote{https://github.com/BuzzFeedNews/2016-10-facebook-fact-check}. These data samples may contain limited scopes of context and styles of writing due to their limited number of categories.

\subsection{Image Datasets}
Most of the existing fake news datasets collect only text data. However, fake news can also come in the form of images. Existing fake image datasets are limited in size and diversity, making dataset research in this area important. Image features supply models with more data that can help immensely to identify fake images and news that have image data. We analyze three traditional fake image datasets that have been published. The Image Manipulation dataset \cite{christlein2012evaluation} contains self-taken manipulated images for image manipulation detection. The PS-Battles dataset \cite{heller2018psBattles} is an image dataset containing manipulated image derivatives from one subreddit. We expand upon the size and scope of the data provided from the same subreddit in the PS-Battles dataset by expanding the size and time range as well as including text data and other metadata. This expanded data makes up only two of 22 sources of data present in our research. The image-verification-corpus~\cite{boididou2018detection}, like ours, contains both text and image data. While it does contain a larger number of samples than other conventional datasets, it still pales in comparison to Fakeddit.

\subsection{Fact-Checking}
Due to the unique aspect of multimodality, Fakeddit can also be applied to the realm of implicit fact-checking. Other existing datasets utilized for fact-checking include FEVER~\cite{thorne-etal-2018-fever} and Fauxtography \cite{zlatkova-etal-2019-fact}. The former consists of altered claims utilized for textual verification. The latter utilizes both text and image data in order to fact-check claims about images. Using both text and image data, researchers can use Fakeddit for verifying truth and proof: utilizing image data as evidence for text truthfulness or using the text data as evidence for image truthfulness.

Compared to other existing datasets, Fakeddit provides a larger breadth of novel features that can be applied in a number of applications: fake news text, image, text+image classification as well as implicit fact-checking. Other data provided, such as comments data, enable more applications.

\section{Fakeddit}
\subsection{Data Collection}
We sourced our dataset from Reddit, a social news and discussion website where users can post submissions on various subreddits. Reddit is one of the top 20 websites in the world by traffic\footnote{https://www.alexa.com/topsites}. Each subreddit has its own theme. For example, `nottheonion' is a subreddit where people post seemingly false stories that are surprisingly true. Active Reddit users are able to upvote, downvote, and submit comments on the submissions.

 Fakeddit consists of over 1 million submissions from 22 different subreddits. The specific subreddits can be found in the Appendix. As depicted in Table \ref{tab:stats}, the samples span over almost a decade and are posted on highly active and popular pages by over 300,000 unique individual users, allowing us to capture a wide variety of perspectives. Having a decade's worth of recent data allows machine learning models to stay attuned to contemporary cultural-linguistic patterns and current events. Our data also varies in its content, because of the array of the chosen subreddits, ranging from political news stories to simple everyday posts by Reddit users. 
 
 Submissions were collected with the pushshift.io API\footnote{https://pushshift.io/} with the earliest submission being from March 19, 2008, and the most recent submission being from October 24, 2019. We gathered the submission title and image, comments made by users who engaged with the submission, as well as other submission metadata including the score, the username of the author, subreddit source, sourced domain, number of comments, and up-vote to down-vote ratio. From the photoshopbattles subreddit, we treated both submission and comment data as submission data. In the photoshopbattles subreddit, users post submissions that contain true images. Other users then manipulate these submission images and post these doctored images as comments on the submission's page. These comments also contain text data that relate or describe the image. We harvest these comments from the photoshopbattles subreddit and treat them as submission data to incorporate in our submission dataset, significantly contributing to the total number of multimodal samples. Approximately 64\% of the samples in our dataset contain both text and images. These multimodal samples are used for our baseline experiments and error analysis. 
 
\begin{table}[t]
\centering
\small
\makebox[\linewidth]{
\begin{tabular}{{l}|{l}}
\toprule
Dataset Statistics \\
\hline
Total samples & 1,063,106 \\
Fake samples & 628,501 \\
True samples & 527,049 \\
Multimodal samples & 682,996 \\
Subreddits & 22 \\
Unique users & 358,504\\
Unique domains & 24,203\\
Timespan & 3/19/2008 - 10/24/2019 \\
Mean words per submission & 8.27 \\
Mean comments per submission & 17.94\\
Vocabulary size & 175,566 \\
Training set size & 878,218 \\
Validation set size & 92,444 \\
Released test set size & 92,444 \\
Unreleased set size & 92,444 \\

\bottomrule
 \end{tabular}
 }
\caption{Fakeddit dataset statistics}\label{tab:stats}
\end{table}
 
% \begin{figure}
%   \centering
%   \includegraphics[width=\linewidth]{images/4gram.pdf}
%   \caption{Type-caption curve of Fakeddit vs. FEVER with 4-gram type. 35,000 samples were randomly chosen from each dataset. samples.}\label{fig:4gram}
% \end{figure}
% \vspace{0.00mm} 
% \begin{figure}
%   \centering
%   \includegraphics[width=\linewidth]{images/textlength.pdf}
%   \caption{}\label{fig:length}
% \end{figure}
% \begin{figure}
%   \centering
%   \includegraphics[width=\linewidth]{images/userposts.pdf}
%   \caption{[] 4-gram analysis of Fakeddit and FEVER. 25,000 samples were randomly chosen from each dataset. On top of having 5 times more total samples than FEVER, Fakeddit also provides significantly more lexical diversity per n samples.}\label{fig:curve}
% \end{figure}
 
 \subsection{Quality Assurance}
 Because our dataset contains over one million samples, it is crucial to make sure that it contains reliable data. To do so, we have several levels of data processing. The first is provided through the subreddit pages. Each subreddit has moderators that ensure submissions pertain to the subreddit theme. The job of these moderators is to remove posts that violate any rules. As a result, the data goes through its first round of refinement. The next stage occurs when we start collecting the data. In this phase, we utilize Reddit's upvote/downvote score feature. This feature is intended to not only signify another user's approval for the post but also indicates that a post does not contribute to the subreddit's theme or is off-topic if it has a low score\footnote{https://www.reddit.com/wiki/reddiquette/}. As such, we filtered out any submissions that had a score of less than 1 to further ensure that our data is credible. We assume that invalid or irrelevant posts within a subreddit would be either removed or down-voted to a score of less than 1. The high popularity of the Reddit website makes this step particularly effective as thousands of individual users can give their opinion of the quality of various submissions.
 
 Our final degree of quality assurance is done manually and occurs after the previous two stages. We randomly sampled 10 posts from each subreddit in order to determine whether the submissions really do pertain to each subreddit's theme. If any of the 10 samples varied from this, we decided to remove the subreddit from our list. As a result, we ended up with 22 subreddits to keep our processed data after this filtering. When labeling our dataset, we labeled each sample according to its subreddit's theme. These labels were determined during the last processing phase, as we were able to look through many samples for each subreddit. Each subreddit is labeled with one 2-way, 3-way, and 6-way label. Lastly, we cleaned the submission title text: we removed all punctuation, numbers, and revealing words such as ``PsBattle'' and ``colorized'' that automatically reveal the subreddit source. For the savedyouaclick subreddit, we removed text following the ``\mybar'' character and classified it as misleading content. We also converted all the text to lowercase. 
 
 As mentioned above, we do not manually label each sample and instead label our samples based on their respective subreddit's theme. By doing this, we employ distant supervision, a commonly used technique, to create our final labels. While this may create some noise within the dataset, we aim to remove this from our pseudo-labeled data. By going through these stages of quality assurance, we can determine that our final dataset is credible and each subreddit's label will accurately identify the posts that it contains. We test this by randomly sampling 150 text-image pairs from our dataset and having two of our researchers individually manually label them for 6-way classification. It is difficult to narrow down each sample to exactly one subcategory, especially for those not working in the journalism industry. We achieve a Cohen's Kappa coefficient~\cite{cohen1960coefficient} of 0.54, showing moderate agreement and that some samples may represent more than one label. While we only provide each sample with one 6-way label, future work can help identify multiple labels for each text-image pair.

\subsection{Labeling}
We provide three labels for each sample, allowing us to train for 2-way, 3-way, and 6-way classification. Having this hierarchy of labels will enable researchers to train for fake news detection at a high level or a more fine-grained one. The 2-way classification determines whether a sample is fake or true. The 3-way classification determines whether a sample is completely true, the sample is fake and contains text that is true (i.e. direct quotes from propaganda posters), or the sample is fake with false text. Our final 6-way classification was created to categorize different types of fake news rather than just doing a simple binary or trinary classification. This can help in pinpointing the degree and variation of fake news for applications that require this type of fine-grained detection. In addition, it will enable researchers to focus on specific types of fake news classification if they desire; for example, focusing on satire only. For the 6-way classification, the first label is true and the other five are defined within the seven types of fake news~\cite{wardle}. Only five types of fake news were chosen as we did not find subreddits with posts aligning with the remaining two types. We provide examples from each class for 6-way classification in Figure \ref{fig:dataset}. The 6-way classification labels are explained below:

\begin{figure}
  \centering
  \includegraphics[width=\linewidth]{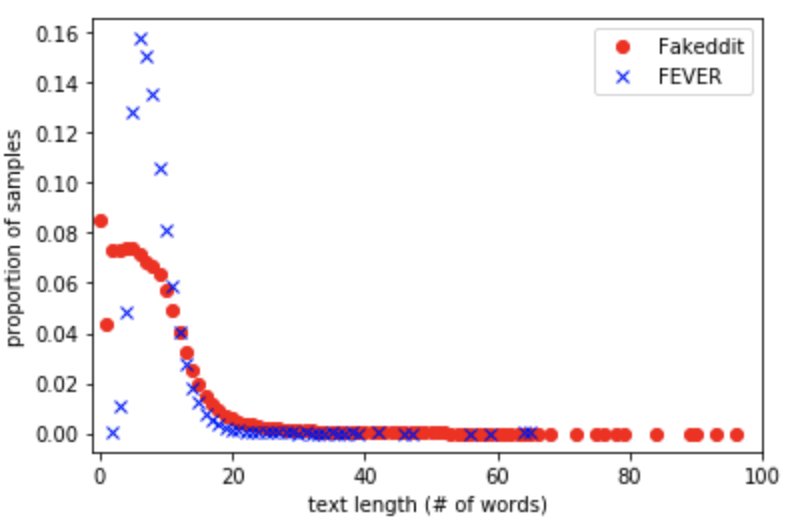}
  \caption{Distributions of word length in Fakeddit and FEVER datasets. We exclude samples that have more than 100 words.}\label{fig:length}
\end{figure}
\begin{figure}
  \centering
  \includegraphics[width=\linewidth]{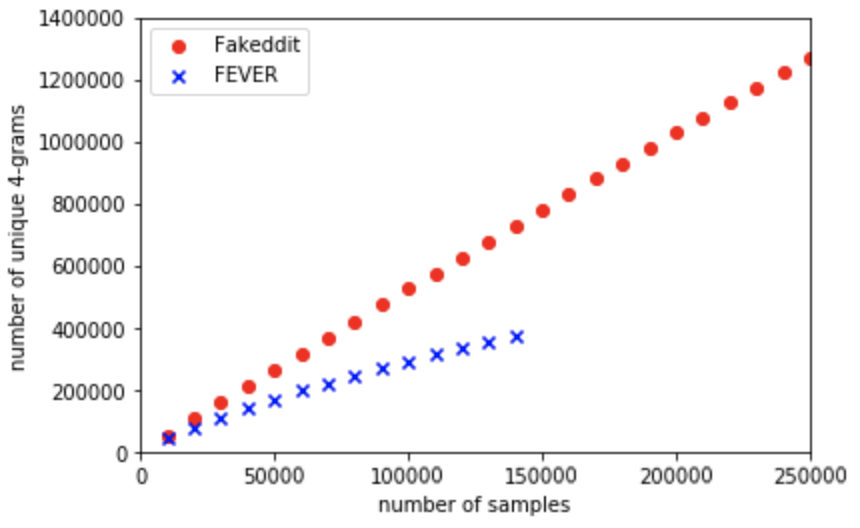}
  \caption{Type-caption curve of Fakeddit vs. FEVER with 4-gram type.}\label{fig:4gram}
\end{figure}

\textbf{True:}
True content is accurate in accordance with fact. Eight of the subreddits fall into this category, such as usnews and mildlyinteresting. The former consists of posts from various news sites. The latter encompasses real photos with accurate captions.

\textbf{Satire/Parody:}
This category consists of content that spins true contemporary content with a satirical tone or information that makes it false. One of the four subreddits that make up this label is theonion, with headlines such as ``Man Lowers Carbon Footprint By Bringing Reusable Bags Every Time He Buys Gas". 

\textbf{Misleading Content:}
This category consists of information that is intentionally manipulated to fool the audience. Our dataset contains three subreddits in this category.

\textbf{Imposter Content:}
This category contains two subreddits, which contain bot-generated content and are trained on a large number of other subreddits. 

\textbf{False Connection:}
 Submission images in this category do not accurately support their text descriptions. We have four subreddits with this label, containing posts of images with captions that do not relate to the true meaning of the image. 
 
\textbf{Manipulated Content:}
Content that has been purposely manipulated through manual photo editing or other forms of alteration. The photoshopbattle subreddit comments (not submissions) make up the entirety of this category. Samples contain doctored derivatives of images from the submissions.

\begin{table}[t]
\centering
\small
\begin{tabular}{{l}|*{4}{c}}
\toprule
Dataset&1-gram&2-gram&3-gram&4-gram\\
\hline
FEVER &40874&179525&315025&387093\\
\hline
Fakeddit &61141&507512&767281&755929 \\

\bottomrule
 \end{tabular}
\caption{Unique n-grams for FEVER and Fakeddit for equal sample size (FEVER's total dataset size). }\label{tab:n-grams}
\end{table}

\begin{figure}
  \centering
  \includegraphics[width=0.9\linewidth]{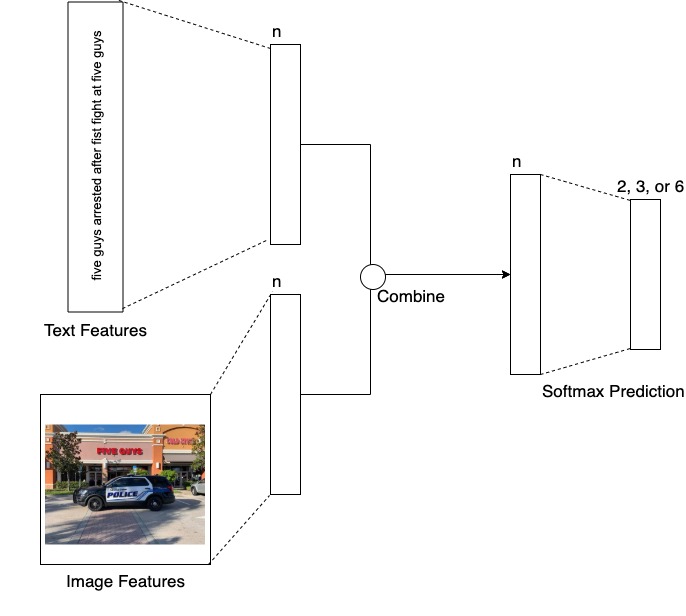}
  \caption{Multimodal model for integrating text and image data for 2, 3, and 6-way classification. \textit{n}, the hidden layer size, is tuned for each model instance through hyperparameter optimization. }\label{fig:model}
\end{figure}

\begin{table*}[t]
\centering
\small
\begin{tabular}{{l}{l}{l}|*{2}{c}|*{2}{c}|*{2}{c}}
\toprule
&&&  \multicolumn{2}{c}{2-way}& \multicolumn{2}{c}{3-way}& \multicolumn{2}{c}{6-way}  \\
\hline
Type & Text & Image & Validation  & Test & Validation  & Test& Validation  & Test \\
\hline
Text & BERT & -- & \textbf{0.8654} & \textbf{0.8644} & \textbf{0.8582} & \textbf{0.8580} & \textbf{0.7696} & \textbf{0.7677} \\
& InferSent & -- & 0.8634 & 0.8631 & 0.8569 & 0.8570 & 0.7652 & 0.7666 \\
\hline
Image & -- & VGG16 & 0.7355 & 0.7376 & 0.7264 & 0.7293 & 0.6462 & 0.6516 \\
&-- & EfficientNet & 0.6115 & 0.6087 & 0.5877 & 0.5828 & 0.4152 & 0.4153 \\
&-- & ResNet50 & \textbf{0.8043} & \textbf{0.8070} & \textbf{0.7966} & \textbf{0.7988} & \textbf{0.7529} & \textbf{0.7549}  \\
\hline
Text+Image & InferSent & VGG16 & 0.8655 & 0.8658 & 0.8618 & 0.8624 & 0.8130 & 0.8130 \\ 
&InferSent & EfficientNet & 0.8328 & 0.8339 & 0.8259 & 0.8256 & 0.7266 & 0.7280 \\
&InferSent & ResNet50 & 0.8888 & 0.8891 & 0.8855 & 0.8863 & 0.8546 & 0.8526 \\
&BERT & VGG16 & 0.8694 & 0.8699 & 0.8644 & 0.8655 & 0.8177 & 0.8208 \\
&BERT & EfficientNet & 0.8334 & 0.8318 & 0.8265 & 0.8255 & 0.7258 & 0.7272 \\
&BERT & ResNet50 & \textbf{0.8929} &\textbf{0.8909} & \textbf{0.8905} & \textbf{0.8890} & \textbf{0.8600} & \textbf{0.8588} \\
\bottomrule
\end{tabular}

\caption{Results on fake news detection for 2, 3, and 6-way classification with combination method of maximum.}\label{tab:results}
\end{table*}

\subsection{Dataset Analysis}
In Table \ref{tab:stats}, we provide an overview of specific statistics pertaining to our dataset such as vocabulary size and number of unique users. We also provide a more in-depth analysis in comparison to another sizable dataset, FEVER. 

First, we choose to examine the word lengths of our text data. Figure \ref{fig:length} shows the proportion of samples per text length for both Fakeddit and FEVER. It can be seen that our dataset contains a higher proportion of longer text starting from word lengths of around 17, while FEVER's captions peak at around 10 words. In addition, while FEVER's peak is very sharp, Fakeddit has a much smaller and more gradual slope. Fakeddit also provides a broader diversity of text lengths, with samples containing almost 100 words. Meanwhile, FEVER's longest text length stops at less than 70 words.

Secondly, we examine the linguistic variety of our dataset by computing the Type-Caption Curve, as defined in \cite{Wang_2019_ICCV}. Figure \ref{fig:4gram} shows these results. Fakeddit provides significantly more lexical diversity. Even though Fakeddit contains more samples than FEVER, the number of unique n-grams contained in similar sized samples are still much higher than those within FEVER. These effects will be magnified as Fakeddit contains more than 5 times more total samples than FEVER. In Table~\ref{tab:n-grams}, we show the number of unique n-grams for both datasets when sampling $n$ samples, where $n$ is equal to FEVER's dataset size. This demonstrates that for all n-gram sizes, our dataset is more lexically diverse than FEVER's for equal sample sizes.

These salient text features - longer text lengths, broad array of text lengths, and higher linguistic variety - highlight Fakeddit's diversity. This diversity can strengthen fake news detection systems by increasing their lexical scope.

\section{Experiments}

\subsection{Fake News Detection}
Multiple methods were employed for text and image feature extraction. We used InferSent~\cite{conneau-EtAl:2017:EMNLP2017} and BERT \cite{devlin-etal-2019-bert} to generate text embeddings for the title of the Reddit submissions. VGG16~\cite{Simonyan15}, EfficientNet~\cite{tan2019efficientnet}, and ResNet50~\cite{he2016deep} were utilized to extract the features of the Reddit submission thumbnails.

We used the InferSent model because it performs very well as a universal sentence embeddings generator. For this model, we loaded a vocabulary of 1 million of the most common words in English and used fastText embeddings \cite{joulin-etal-2017-bag}. We obtained encoded sentence features of length 4096 for each submission title using InferSent. 

\begin{table*}[t]
\centering
\small
\begin{tabular}{{l}|*{2}{c}|*{2}{c}|*{2}{c}}
\toprule
&  \multicolumn{2}{c}{2-way}& \multicolumn{2}{c}{3-way}& \multicolumn{2}{c}{6-way}  \\
\hline
Combination Methods  & Validation  & Test & Validation  & Test& Validation  & Test \\
\hline
Add & 0.8551 & 0.8551 & 0.8509 & 0.8505 & 0.8206 & 0.8235 \\
Concatenate & 0.8564 & 0.8568 & 0.8531 & 0.8530 & 0.8237 & 0.8249 \\
Maximum & \textbf{0.8929} &\textbf{0.8909} & \textbf{0.8905} & \textbf{0.8890} & \textbf{0.8600} & \textbf{0.8588} \\
Average & 0.8554 & 0.8561 & 0.8512 & 0.8518 & 0.8229 & 0.8242 \\
\bottomrule
 \end{tabular}
\caption{Results on different multi-modal combinations for BERT + ResNet50}\label{tab:combine}
\end{table*}

% \begin{table*}[!htbp]
% \centering
% \small
% \begin{tabular}{{l}|*{2}{c}|*{2}{c}|*{2}{c}|*{2}{c}}
% \toprule
% &  \multicolumn{2}{c}{2-way}& \multicolumn{2}{c}{3-way}& \multicolumn{2}{c}{6-way} & \multicolumn{2}{c}{22-way}  \\
% \hline
% Features & Validation  & Test & Validation  & Test& Validation  & Test & Validation  & Test \\
% \hline
% Text only & [] & ? & ? & ? & ? & ? & ? & ? \\
% Image only & ? & ? & ? & ? & ? & ? & ? & ? \\
% Multimodal (Max) & ? & ? & ? & ? & ? & ? & ? & ? \\
% \bottomrule
%  \end{tabular}
% \caption{Results on control tasks. BERT and ResNet50 features were utilized. }\label{tab:control-tasks}
% \end{table*}

In addition, we used the BERT model. BERT achieves state-of-the-art results on many classification tasks, including Q\&A and named entity recognition. To obtain fixed-length BERT embedding vectors, we used the bert-as-service\cite{xiao2018bertservice} tool, to map variable-length text/sentences into a 768 element array for each Reddit submission title. For our experiments, we utilized the pretrained BERT-Large, Uncased model. 

We employed VGG16, ResNet50, and EfficientNet models for encoding images. VGG16 and ResNet50 are widely used by many researchers, while EfficientNet is a relatively newer model. For EfficientNet, we used variation: B4. This was chosen as it is comparable to ResNet50 in terms of FLOP count. For the image models, we preloaded weights of models trained on ImageNet and included the top layer and used the penultimate layer for feature extraction. 

\subsection{Experiment Settings}

As mentioned in section 3.2, the text was cleaned thoroughly through a series of steps. We also prepared the images by constraining the sizes of the images to match the input size of the image models. We applied necessary image preprocessing required for the image models.  

For our experiments, we excluded submissions that have either text or image data missing. We performed 2-way, 3-way, and 6-way classification for each of the three types of inputs: image only, text only, and multimodal (text and image). As in Figure~\ref{fig:model}, when combining the features in multimodal classification, we first condensed them into n-element vectors through a trainable dense layer and then merged them through four different methods: add, concatenate, maximum, average. These features were then passed through a fully connected softmax predictor. For all experiments, we tuned the hyperparameters on the validation dataset using the keras-tuner tool\footnote{https://github.com/keras-team/keras-tuner}. Specifically, we employed the Hyperband tuner~\cite{li2018hyperband} to find optimal hyperparameters for the hidden layer size and learning rates. The hyperparameters are tuned on the validation set. We varied the number of units in the hidden layer from 32 to 256 with increments of 32. For the optimizer, we used Adam~\cite{kingma2014adam} and tested three learning rate values: 1e-2, 1e-3, 1e-4. For the multimodal model, the unit size hyperparameter affected the sizes of the 3 layers simultaneously: the 2 layers that are combined and the layer that is the result of the combination. For non-multimodal models, we utilized a single size-tunable hidden layer, followed by a softmax predictor. For each model, we specified a maximum of 20 epochs and an early stopping callback to halt training if the validation accuracy decreased.

% \subsection{Control Tasks}
% In addition, we verified the models’ abilities to discriminate different categories of fake news. Similar to the approach presented in \cite{hewtt} [: They have an ACL Paper, convert to ACL paper citation], we conducted control tasks to assess the models’ selectivity and ability to generalize. We labeled each subreddit in our dataset a random label from 1 - 6 regardless of their true 6-way category. For example, all samples from the subreddit ‘theonion’ and ‘nottheonion’ could receive the same label such as False Connection even though they are from different fake news categories. We conducted training with these altered labels to test whether the system was doing mere subreddit categorization or legitimate fake news categorization. We conducted similar experiments for 2-way and 3-way classification as well, assigning random labels for each subreddit based on the type of classification. 

\subsection{Results}
The results are shown in Tables \ref{tab:results} and \ref{tab:combine}. For image and multimodal classification, ResNet50 performed the best followed by VGG16 and EfficientNet. In addition, BERT achieved better results than InferSent for multimodal classification. Multimodal features performed the best, followed by text-only, and image-only. Thus, image and text multimodality present in our dataset significantly improves fake news detection. The ``maximum'' method to merge image and text features yielded the highest accuracy. Overall, the multimodal model that combined BERT text features and ResNet50 image features through the maximum method performed most optimally. The best 6-way classification model parameters were: hidden layer sizes of 224 units, 1e-4 learning rate, trained over 20 epochs. 

% In addition, we compared the results from these control tasks shown in \ref{tab:control-tasks} against those in \ref{tab:results} to determine the model's ability to generalize fake news. For all types of classification, we found that baseline model experiments outperformed control tasks. Thus, we confirm the models were not merely classifying subreddits, but were rather understanding the semantics of fake news in both binary and fine-grained classification. Thus, we establish the high quality of our dataset. 

\newcolumntype{L}{>{\centering\arraybackslash}m{3cm}}

\begin{table*}[t]
\centering
\small
\makebox[\linewidth]{
\begin{tabular}{{l}{l}|{l}|{l}|{l}}
\toprule
Text &Image  & Predicted Label  & 
Gold Label & PM(\%)  \\
\hline
 \multicolumn{1}{m{3cm}}{volcanic eruption in bali last night}& \includegraphics[width=0.1\textwidth,
                             margin=0pt 1ex 0pt 1ex,valign=m]{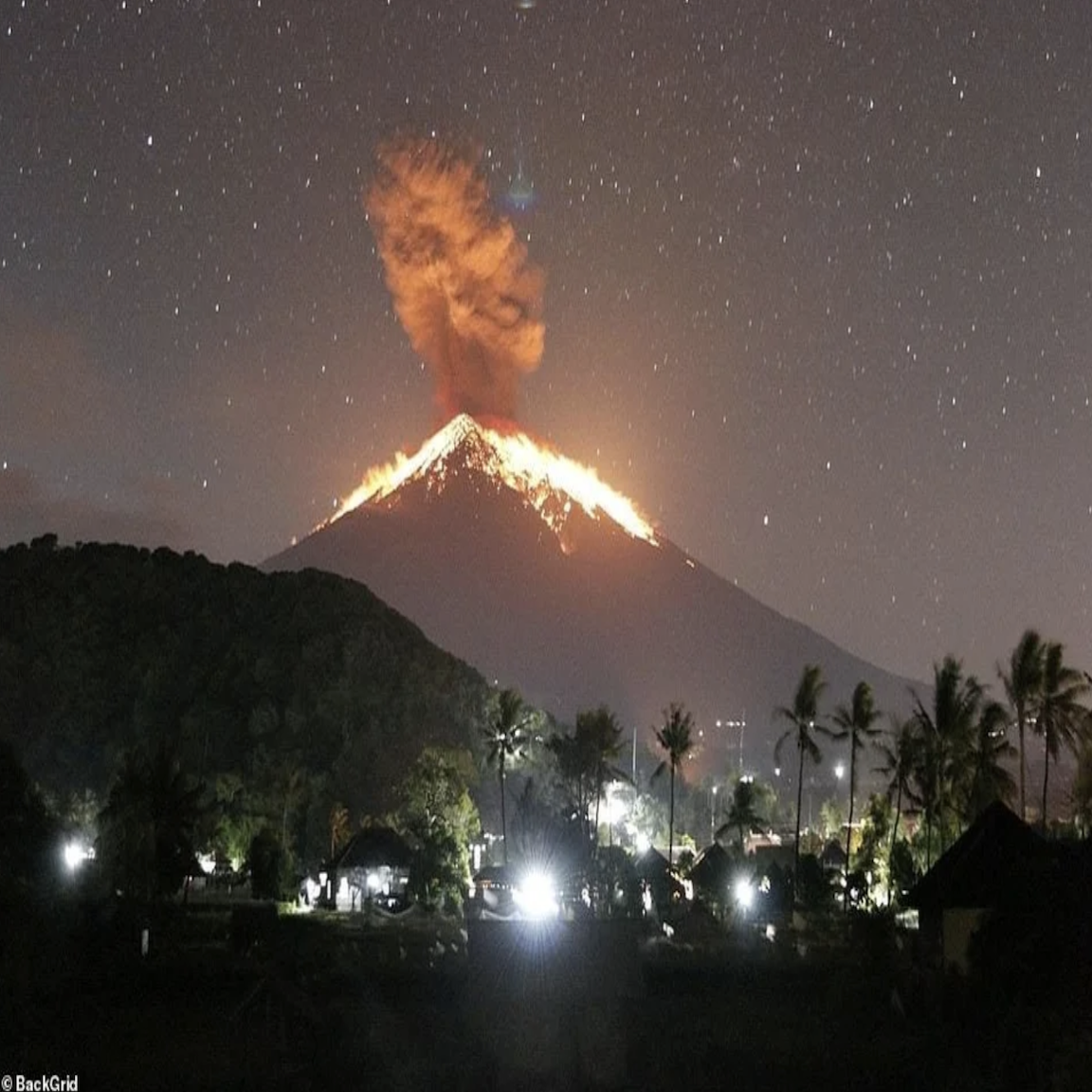}  & False Connection  & True&17.9\\
\hline
\multicolumn{1}{m{3cm}}{nascar race stops to wait for family of ducks to pass} & \cincludegraphics[width=0.1\textwidth,
                             margin=0pt 1ex 0pt 1ex,valign=m]{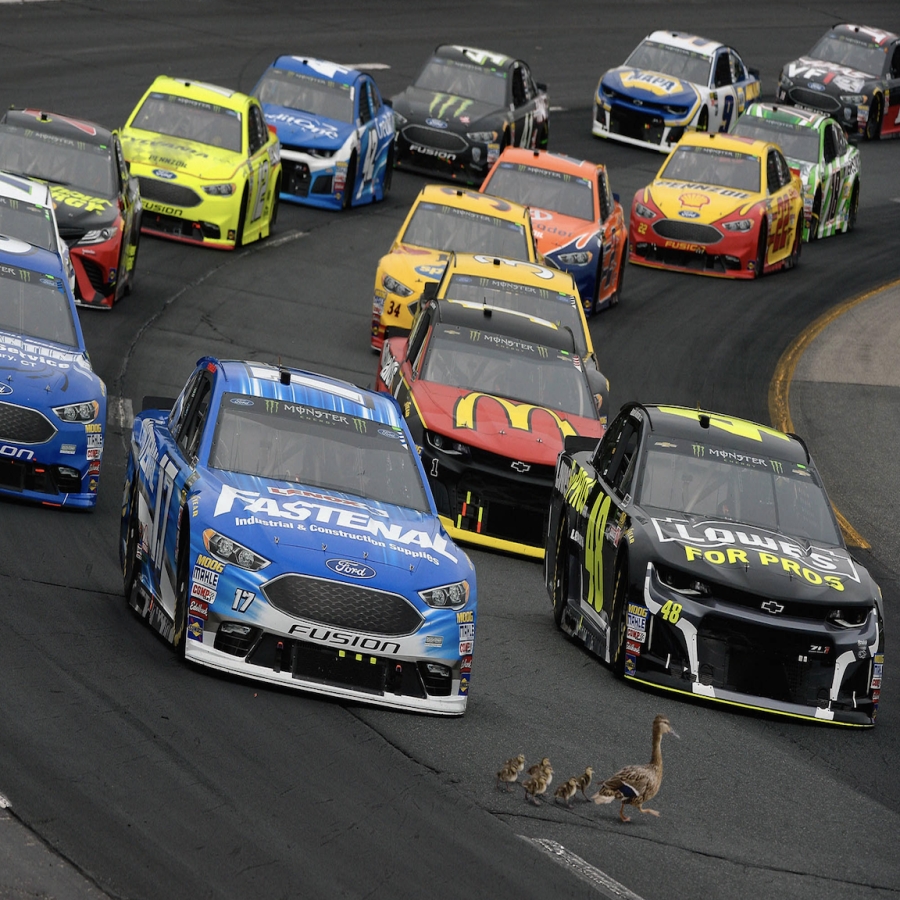}  & True  & Satire&32.8\\
\hline
\multicolumn{1}{m{3cm}}{cars race towards nuclear explosion} & \includegraphics[width=0.1\textwidth,
                             margin=0pt 1ex 0pt 1ex,valign=m]{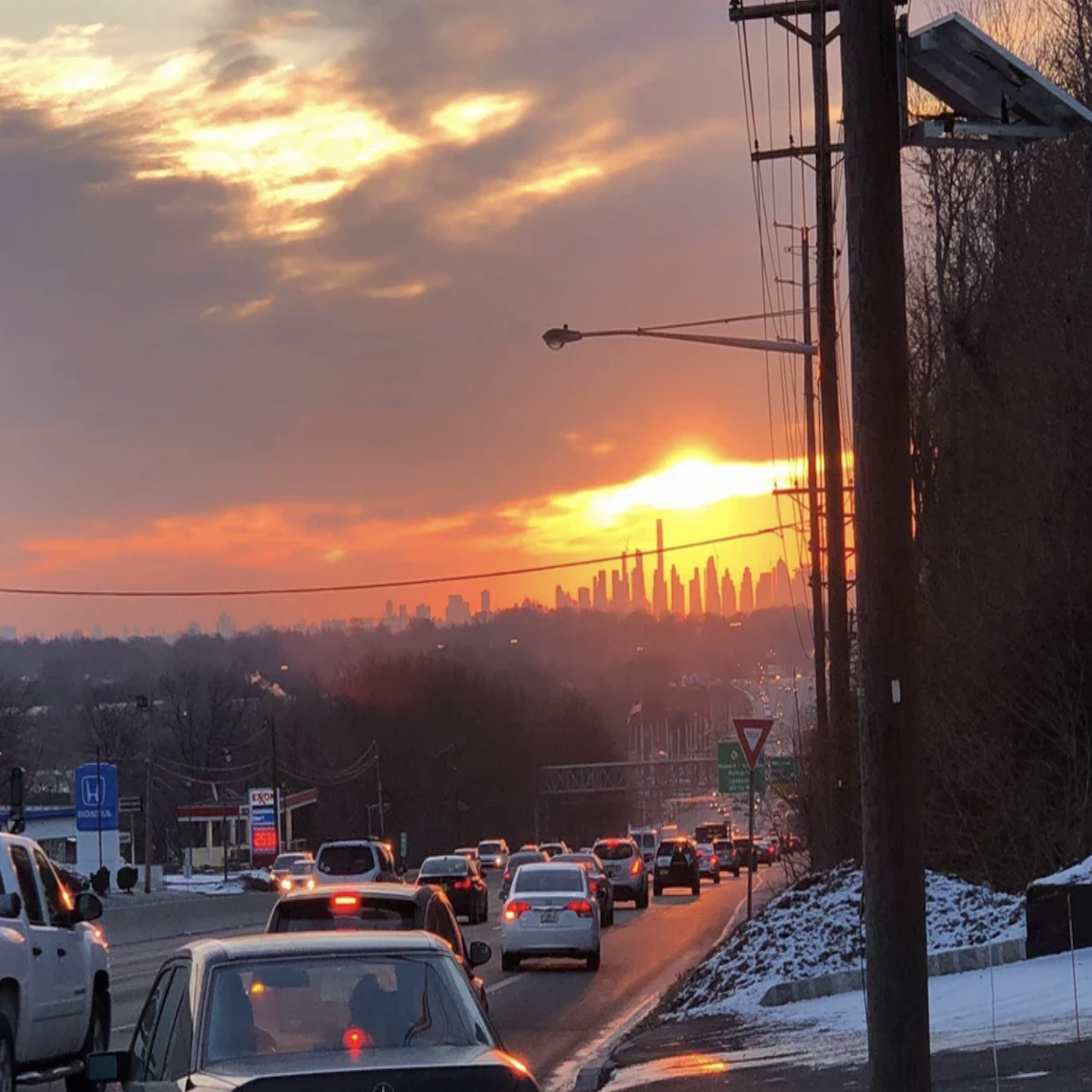}  & True &False Connection&17.8\\
\hline
\multicolumn{1}{m{3cm}}{bear experiences getting hit in the cinema rule, your child again} & \cincludegraphics[width=0.1\textwidth,
                             margin=0pt 1ex 0pt 1ex,valign=m]{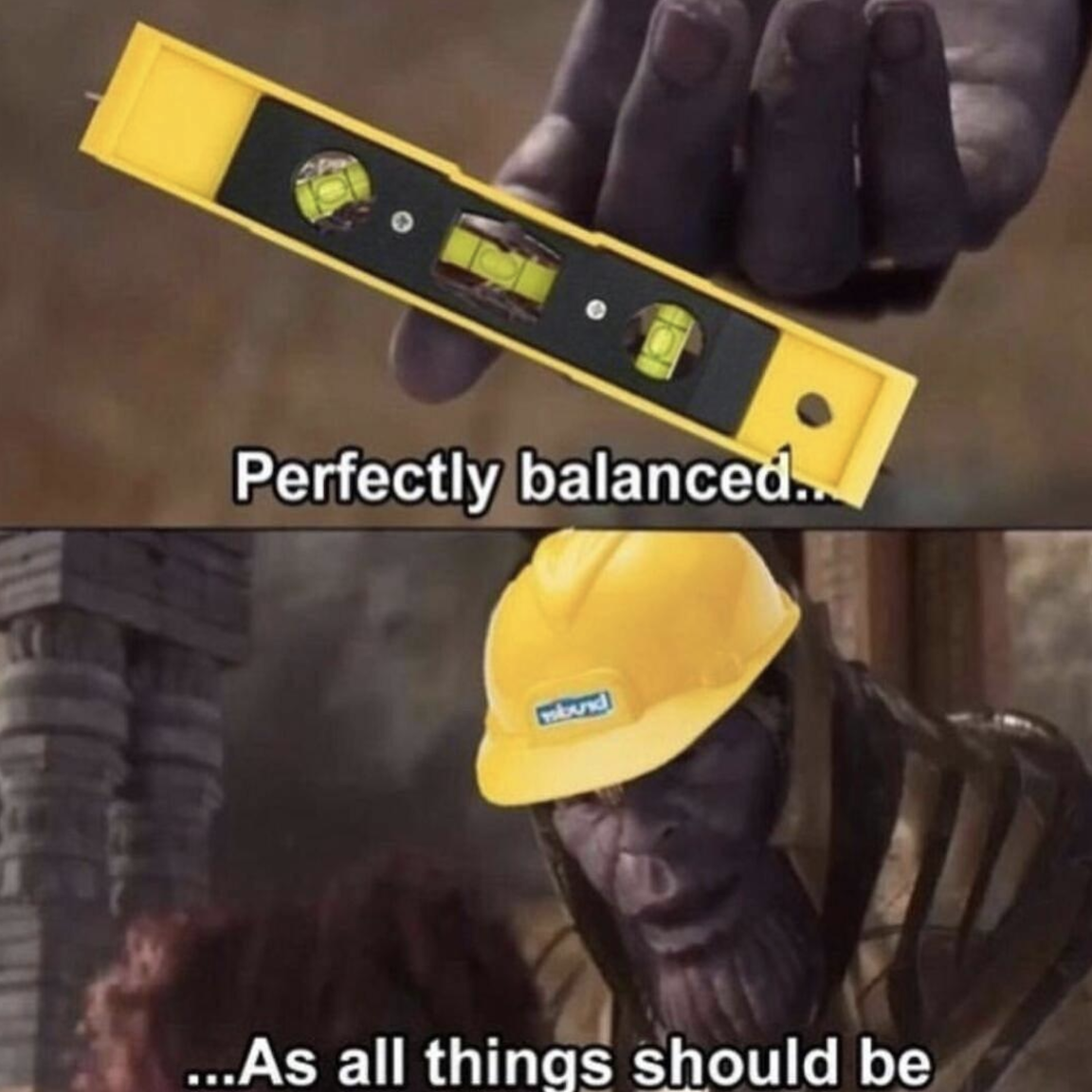}  & Satire &Imposter Content&55.7\\
\hline
\multicolumn{1}{m{3cm}}{three corgis larping at the beach} & \cincludegraphics[width=0.1\textwidth,
                             margin=0pt 1ex 0pt 1ex,valign=m]{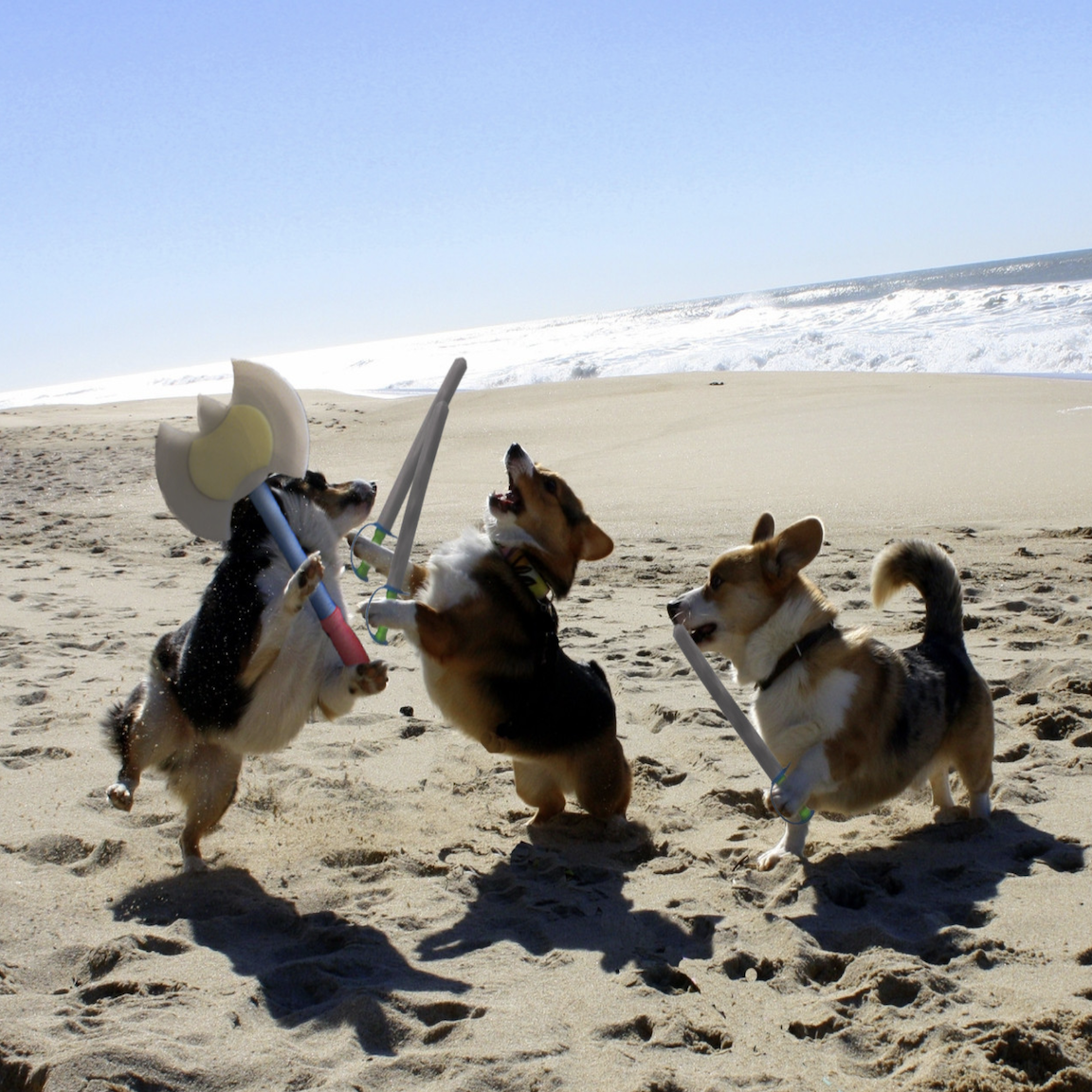}  & True & Manipulated Content&3.3\\
\hline
\multicolumn{1}{m{3cm}}{mighty britain getting tied down in south africa during boer bar circa} & \cincludegraphics[width=0.1\textwidth,
                             margin=0pt 1ex 0pt 1ex,valign=m]{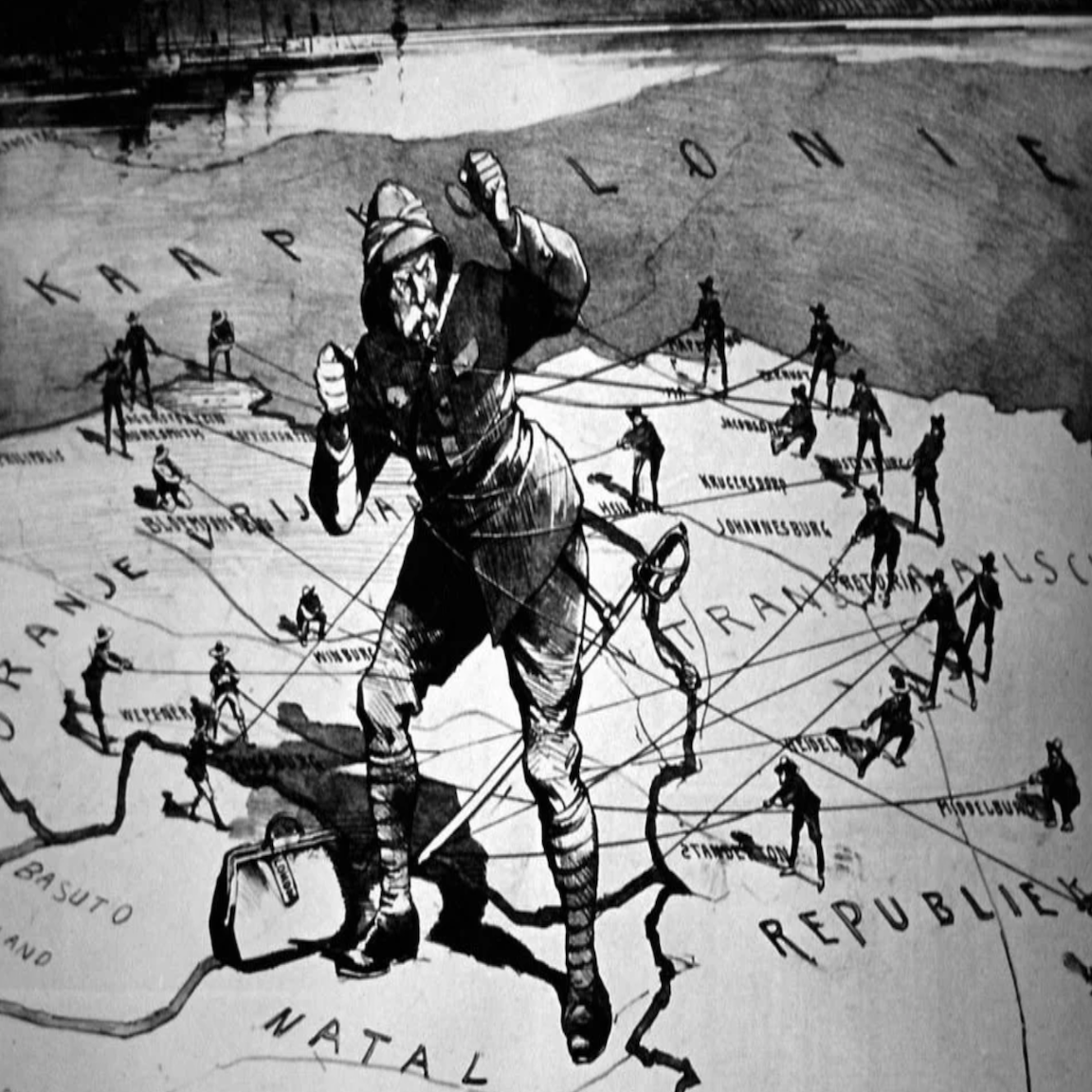}  & False Connection  & Misleading Content&16.9\\
\bottomrule
 \end{tabular}
 }
\caption{Classification errors on the BERT+ResNet50 model for 6-way classification. PM: Proportion of samples misclassified within each Gold label.}\label{tab:error}
\end{table*}

\section{Error Analysis}
We conduct an error analysis on our 6-way detection model by examining samples from the test set that the model predicted incorrectly. A subset of these samples is shown in Table \ref{tab:error}. Firstly, the model had the most difficult time identifying imposter content. This category contains subreddits that contain machine-generated samples. Recent advances in machine learning such as Grover~\cite{zellers2019grover}, a model that produces realistic-looking machine-generated news articles, have allowed machines to automatically generation human-like material. Our model has a relatively difficult time identifying these samples. The second category the model had the poorest performance on was satire samples. The model may have a difficult time identifying satire because creators of satire tend to focus on creating content that seems similar to real news if one does not have a sufficient level of contextual knowledge. Classifying the data into these two categories (imposter content and satire) are complex challenges, and our baseline results show that there is significant room for improvement in these areas. On the other hand, the model was able to correctly classify almost all manipulated content samples. We also found that misclassified samples were frequently categorized as being true. This can be attributed to the relative size of true samples in the 6-way classification. While we have comparable sizes of fake and true samples for 2-way classification, 6-way breaks down the fake news into more fine-grained classes. As a result, the model trains on a higher number of true samples and may be inclined to predict this label. 

% Even though our model trains on both fake and true news, classifying the data into these two categories is still a complex challenge and shows that there is room for improvement in these areas.
% Adding the more fine-grained labels may also be confusing the model, therefore lowering the accuracy in fake vs. true detection. In addition, this highlights our main objective for creating this dataset: the difficulty of classifying fake vs. true news.

\begin{table*}[ht]
\centering
\small
\makebox[\linewidth]{
\begin{tabular}{{l}|{l}|{l}}
\toprule
Subreddit &6-Way Label & URL   \\
\hline
photoshopbattles submissions & True & https://www.reddit.com/r/photoshopbattles \\
\hline
nottheonion & True & https://www.reddit.com/r/nottheonion \\
\hline
neutralnews & True & https://www.reddit.com/r/neutralnews \\
\hline
pic & True & https://www.reddit.com/r/pic \\
\hline
usanews & True & https://www.reddit.com/r/usanews \\
\hline
upliftingnews & True & https://www.reddit.com/r/upliftingnews \\
\hline
mildlyinteresting & True & https://www.reddit.com/r/mildlyinteresting \\
\hline
usnews & True & https://www.reddit.com/r/usnews \\
\hline
fakealbumcovers & Satire & https://www.reddit.com/r/fakealbumcovers \\
\hline
satire & Satire & https://www.reddit.com/r/satire \\
\hline
waterfordwhispersnews & Satire & https://www.reddit.com/r/waterfordwhispersnews \\
\hline
theonion & Satire & https://www.reddit.com/r/theonion \\
\hline
propagandaposters & Misleading Content & https://www.reddit.com/r/propagandaposters \\
\hline
fakefacts & Misleading Content & https://www.reddit.com/r/fakefacts \\
\hline
savedyouaclick & Misleading Content & https://www.reddit.com/r/savedyouaclick \\
\hline
misleadingthumbnails & False Connection & https://www.reddit.com/r/misleadingthumbnails \\
\hline
confusing\_perspective & False Connection & https://www.reddit.com/r/confusing\_perspective \\
\hline
pareidolia & False Connection & https://www.reddit.com/r/pareidolia \\
\hline
fakehistoryporn & False Connection & https://www.reddit.com/r/fakehistoryporn \\
\hline
subredditsimulator & Imposter Content & https://www.reddit.com/r/subredditsimulator \\
\hline
subsimulatorgpt2 & Imposter Content & https://www.reddit.com/r/subsimulatorgpt2 \\
\hline
photoshopbattles comments & Manipulated Content & https://www.reddit.com/r/photoshopbattles \\

\bottomrule
 \end{tabular}
 }
\caption{List of subreddits used in Fakeddit. }
\label{tab:subred}
\end{table*}

\section{Conclusion}
In this paper, we presented a novel dataset for fake news research, Fakeddit. Compared to previous datasets, Fakeddit provides a large number of multimodal samples with multiple labels for various levels of fine-grained classification. We conducted several experiments with multiple baseline models and performed an error analysis on our results, highlighting the importance of large scale multimodality unique to Fakeddit and demonstrating that there is still significant room for improvement in fine-grained fake news detection. Our dataset has wide-ranging practicalities in fake news research and other research areas. Although we do not utilize submission metadata and comments made by users on the submissions, we anticipate that these additional multimodal features will be useful for further fake news research. For example, future research can look into tracking a user's credibility through using the metadata and comment data provided and incorporating video data as another multimedia source. Implicit fact-checking research with an emphasis on image-caption verification can also be conducted using our dataset's unique multimodality aspect. We hope that our dataset can be used to advance efforts to combat the ever-growing rampant spread of disinformation in today's society.

\section*{Acknowledgments}
We would like to acknowledge Facebook for the Online Safety Benchmark Award. The authors are solely
responsible for the contents of the paper, and the opinions expressed in this publication do not reflect those of the funding agencies.

\section{Bibliographical References}
\bibliographystyle{lrec}
\bibliography{acl2020}

%\clearpage
\newpage

\section*{Appendix}

We show the list of subreddits in Table~\ref{tab:subred}.

% %\section{Language Resource References}
% %\label{lr:ref}
% %\bibliographystylelanguageresource{lrec}
% %\bibliographylanguageresource{lrec2020W-xample}

\end{document}